\documentclass[12pt]{article}

\usepackage{framed,multirow}

\usepackage{url}
\usepackage{xcolor}
\definecolor{newcolor}{rgb}{.8,.349,.1}

\usepackage{graphicx}
\usepackage{amsmath} 
\usepackage{amssymb}

\begin{document}

\thispagestyle{empty}

\title{Crossing the Road Without Traffic Lights:\\ An Android-based Safety Device}

\author{Adi Perry $~~~$ Dor Verbin $~~~$ Nahum Kiryati\\[0.1cm]
School of Electrical Engineering\\ 
Tel Aviv University\\ 
Tel Aviv 69978, Israel}

\date{}

\maketitle

\begin{abstract}
In the absence of pedestrian crossing lights, finding a safe moment to cross the road is often
hazardous and challenging, especially for people with visual impairments.
We present a reliable low-cost solution, an Android device attached 
to a traffic sign or lighting pole near the crossing, indicating whether it is safe to cross the road.
The indication can be by sound, display, vibration, and various communication modalities provided by 
the Android device. The integral system camera is aimed at approaching traffic. 
Optical flow is computed from the incoming video stream, and projected onto an influx map,
automatically acquired during a brief training period. The crossing safety is determined 
based on a 1-dimensional temporal signal derived from the projection.
We implemented the complete system on a Samsung Galaxy K-zoom Android smartphone,
and obtained real-time operation. The system achieves promising experimental results,
providing pedestrians with sufficiently early warning of approaching vehicles.
The system can serve as a stand-alone safety device, that can be installed where pedestrian 
crossing lights are ruled out. Requiring no dedicated infrastructure, it can be powered by a solar panel and 
remotely maintained via the cellular network.     \\[0.2cm]
{\bf Keywords:} 
pedestrian crossing, traffic analysis, optical flow, blind \& visually impaired,
resource-limited computer vision, Android device
\end{abstract}

\section{Introduction}

Safe road crossing requires conflict-free sharing of road surface.
Pedestrian crossing lights bring traffic to a halt to ensure secure crossing.
Other right of way regulations, such as zebra-crossings, are less reliable.
Drivers do not consistently yield to pedestrians~\cite{Wei2015,Schroeder2011}, 
even to blind persons holding a white cane. Thus, for safe crossing in the absence of crossing lights, 
pedestrians need to identify occasional traffic gaps~\cite{Emerson08}.

The duration of a traffic gap allowing safe crossing is determined by
the crossing time. US data~\cite{AASTHO-04} indicates a representative crossing 
speed of about 1.2m/s\, and urban road widths of 5.5m (one-way single lane)
or 8.5m (two-way or two-lane one-way) with shoulders. The corresponding crossing 
times are 4.5s and 7s respectively. The necessary traffic gap, equivalent to the needed advance
warning of an approaching vehicle, must be at least that long.
 
Recognizing an approaching vehicle 7s ahead of contact time is possible for people
with normal or corrected vision and solid cognitive function.
For blind and visually impaired people, as well as for elderly persons, children~\cite{Morrongiello2016} and others,
sufficiently early recognition of an approaching vehicle can be implausible.   
Blind people are trained to rely on hearing and follow the ``cross when quiet'' 
strategy~\cite{Sauerburger99}. Auditory-based performance is less than ideal~\cite{Hassan2012,Barton2012}
and further difficulties arise where background urban noise 
masks the low noise emitted by a modern car~\cite{Emerson2012}, 
or when the pedestrian has less than perfect hearing. 
  
lmage and video processing methods to assist visually handicapped people
were reviewed in~\cite{Pun07,Assistive2008,Leo2016}. Many techniques convert visual data
to auditory or haptic signals. These include systems to locate street crossings~\cite{Shioyama04,Ivachenko08,Murali2013}.
Converting traffic light status to an auditory or other indication was considered
in~\cite{APS03,Bohonos07,Ivanchenko2010}. Sensor-rich ``smart canes'' have been 
developed~\cite{Kay74,Fallon08,Buchs14} for close-range obstacle detection
and path planning. However, the problem of road crossing at unsignalized crosswalks has remained open. 
A noteworthy exception is the work of~\cite{Baker2005},
aimed at providing a robotic system with street-crossing capability.

We recently presented~\cite{Perry2014} a system positioned near 
a road crossing, alerting
pedestrians of approaching traffic and detecting traffic-gaps
suitable for safe crossing. The experimental results reported in~\cite{Perry2014}
were obtained  using a laptop computer connected to an external
USB camera with an add-on narrow field of view lens.
The major open question concerned the feasibility of implementation 
on a platform powerful enough to achieve real-time operation and
meet the optical requirements, cheap enough to allow   
widespread installation, and power-frugal
to permit stand-alone solar-powered operation. 
This paper expands~\cite{Perry2014}
by satisfying these requirements using an off-the-shelf
Android device, and by demonstrating successful autonomous operation 
at actual road crossings\footnote{Video: \url{www.eng.tau.ac.il/~nk/crosswalk-safety.html}}.

\section{Stationary Location-Specific Solution}

Consider a vehicle approaching a crosswalk at 50km/h. Detecting this vehicle 7s in advance,
implies detection at about 100m away from the crosswalk. At that distance, the viewing angle of 
the front of a typical car is about $1^o$ horizontally and slightly less vertically. In many suburban
and country roads, speeds up to 100km/h are not unusual, corresponding to roughly $0.5^o$ at the 
required detection distance.

Assuming a straight road towards the crosswalk, the tiny-looking approaching vehicle
appears near the vanishing point in the picture plane. 
Reliable detection requires predetermination of the point of first appearance, 
and narrow field of view monitoring centered at that point. With some exaggeration, 
this can be illustrated as watching the scene through a straw.   
One might wish for a smartphone application warning the holder of approaching vehicles.
However, typical mobile phones have wide field of view lenses.
Worse, a hand-held device would require careful pointing towards the point of first appearance, 
a challenge for anyone, and a formidable undertaking for people with visual, cognitive or motoric impairments.

We suggest a stationary and location-specific solution, where the system is attached, for example,
to a traffic sign or a lighting pole, facing the approaching traffic.  
The absence of substantial ego-motion improves robustness, facilitates training and adaptation,
and reduces the computational requirements, thus minimizing cost and power consumption. 
Beyond its technical advantages, the location-specific approach does not require a personal device,
and is thus accessible to all members of society.   
 
\begin{figure}[t!]
\begin{center}
\includegraphics[width=30pc]{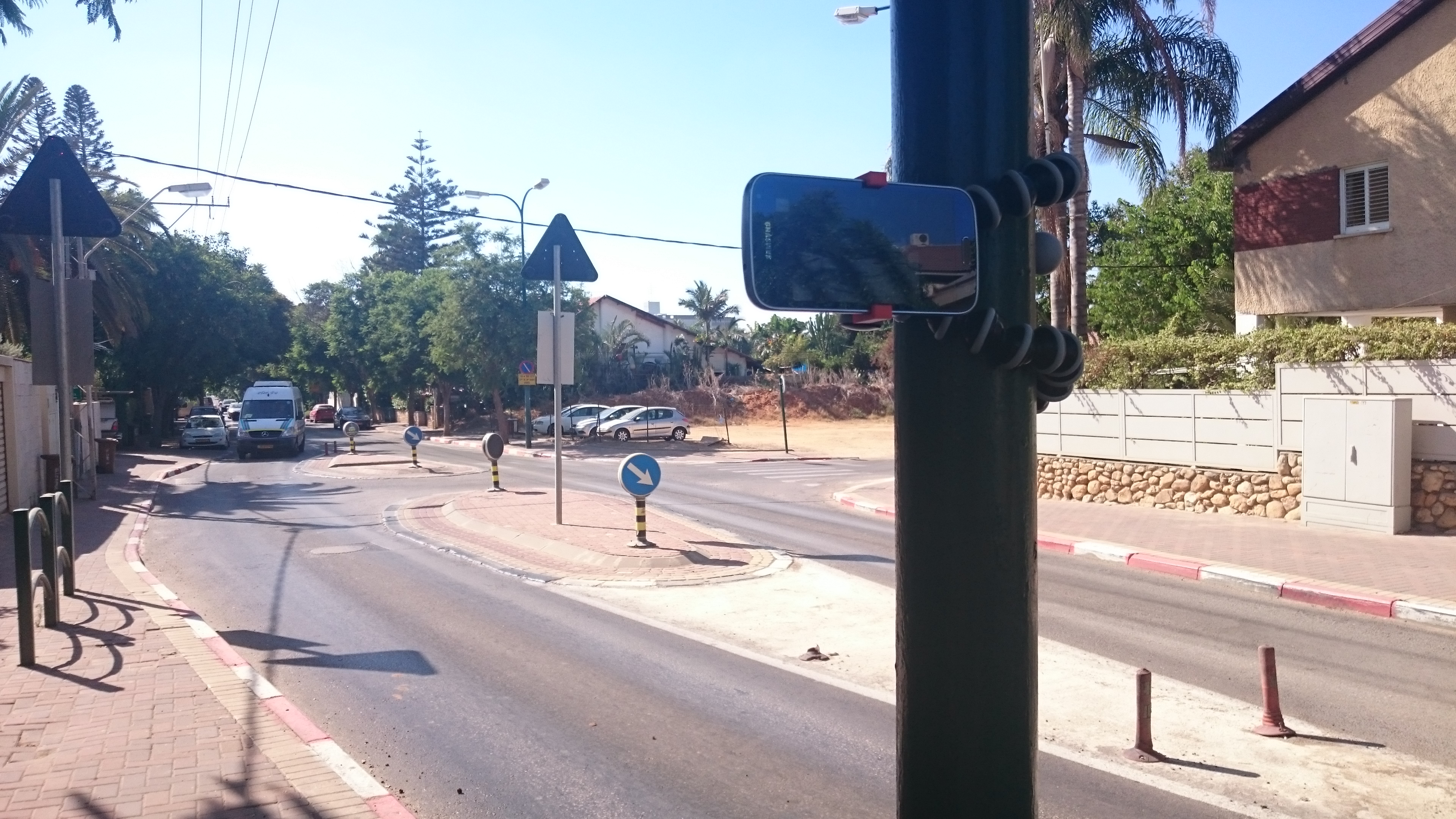}
\end{center}
\caption{The system attached to a lighting pole near a road crossing. The camera faces the approaching traffic. 
\label{fig:system}}
\end{figure}

The essential system elements are a video camera with a narrow field of view lens, a processing unit,
and a user interface. Optional elements include a compact solar panel and a rechargeable battery, enabling
off-grid operation, and a communication unit (cellular, WiFi, etc.), allowing remote maintenance 
as well as cooperation with nearby instances of the system, e.g. in two-way multi-lane road crossings.
As discussed in the sequel, we successfully implemented the system on the Samsung Galaxy K-Zoom
smartphone, see Fig.~\ref{fig:system}. 

Indication of incoming traffic, or lack thereof, can be delivered
to the user as an audio, tactile or visual signal, or via a local networking
interface (such as Bluetooth or WLAN) to a smartphone or a dedicated 
receiver.

\section{Principles of Operation}

Vehicles approaching a crosswalk emerge as tiny blobs at the 
point of first appearance. A key design choice has been to provide the pedestrian 
with the earliest possible warning of an approaching vehicle, rather than the latest 
warning that would allow crossing at the assumed walking speed. This implies that 
the task at hand is early detection, rather than precise estimation of the time to contact 
(TTC)~\cite{Lee76,Horn07}. 

We transform the incoming video to a scalar temporal signal, referred to as~\emph{Activity},
quantifying the overall risk-inducing motion in the scene. 
Vehicles approaching the camera lead to pulses in the Activity signal, so that
substantial activity ascent reflects an approaching vehicle.
Timely detection of incoming traffic, with sparse false alarms, calls for reliable
separation between meaningful Activity ascent and random-like fluctuations.

\subsection{Activity Estimation}

Motion in the scene leads to optical flow. 
Parts of the optical flow field may correspond to vehicles approaching the camera.
Other optical flow elements are due to harmless phenomena, such as 
traffic in other directions, crossing pedestrians, animals on the road, moving leaves 
and slight camera vibrations.
The suggested activity signal $A_n$, where $n$ is the temporal index, is computed as the 
projection of the optical flow field 
 $\vec{u}_n(x,y)$ onto an {\em influx map}. The influx map $\vec{m}(x,y)$ is a vector field,
representing the location-dependent typical course of traffic approaching the camera.
The effective support of the influx map largely corresponds to incoming traffic-lanes.  
Mathematically, 
\begin{equation}
A_n=\sum_{x,y}\vec{m}(x,y)\cdot\vec{u}_n(x,y).
\label{eq:activity}
\end{equation}
The scalar activity signal quantifies the overall hazardous traffic in the scene,
sequentially obtained as the dot product between the current optical flow field and the influx map.

\begin{figure}[t!]
\begin{center}
\includegraphics[width=30pc]{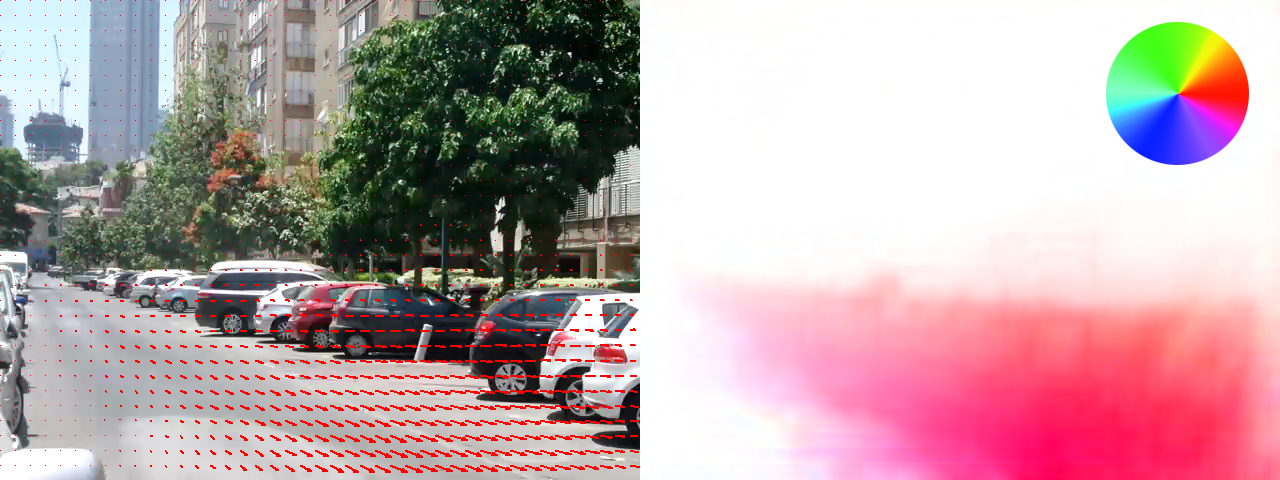}
\end{center}
\caption{Influx map, computed by 
averaging the optical flow field over a training period.
Left: Arrow-style vector display.
Right: Color coded vector representation, corresponding to the color-key shown top-right.
The motion direction in each pixel is represented by hue,
its norm by saturation. 
\label{fig:Projection-Map-Illusration-2}
}
\end{figure}

Assuming a one-way street, the influx map can be obtained
by temporal averaging the optical flow field over a training period, 
as illustrated in Fig.~\ref{fig:Projection-Map-Illusration-2}.
Temporal averaging cancels
randomly oriented motion elements due to vibrations, wind 
and similar effects, while maintaining the salient hazardous traffic motion patterns
in the road area.

\begin{figure}[bt]
\begin{center}
\includegraphics[width=20pc]{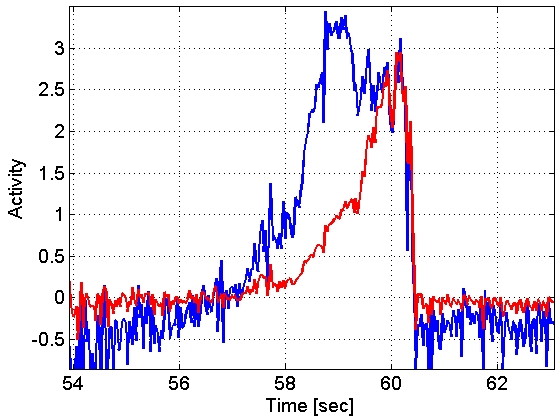}
\end{center}
\caption{Activity signal associated with an approaching vehicle, 
with (blue) and without (red) intensification of the influx map near the 
point of first appearance.}
\label{fig:Activity-Illustration}
\end{figure}

The influx-map vectors can be intensified to emphasize motion
near the point of first appearance, thus increasing the advance warning time,
see  Fig.~\ref{fig:Activity-Illustration}.
Note that the point of first appearance can be identified during training,
e.g. by back-tracking the influx-map vectors to the source.
Minor adjustment of the influx map computation procedure is necessary for 
handing two-way roads, where consistent outbound traffic is also expected. 

We have tacitly presented optical flow and influx map computation in {\em dense} form,
meaning that their spatial resolution is uniform and commensurate with the resolution of the video 
frame. However, {\em sparse} computation of the optical flow and influx map, at spatially variant density,
is beneficial with regard to both performance and computational cost. Foveated computation
of  Eq.~\ref{eq:activity}, dense near the point of first appearance and sparse elsewhere, 
has the desirable effect of emphasizing motion in that critical region.

\subsection{Detecting Approaching Vehicles}

Based on the raw Activity signal, we wish to generate a binary indication,
suggesting whether risk-inducing traffic is approaching (`TRAFFIC' state),
or a sufficient traffic gap allows safe crossing ('GAP' state).  

As a vehicle approaches the camera, the peak of the corresponding  
Activity pulse is usually evident. However, {\em early} warning requires detection
at an early stage of ascent, where the signal to noise ratio is poor.

Reliable, timely detection cannot be achieved by direct thresholding of the Activity signal.
The signal is therefore evaluated within a sliding temporal window of $N$ samples. 
This improves the effective signal to noise ratio, consequently increasing the 
detection probability and reducing the false alarm rate. However, sliding window 
computation inherently introduces detection latency.
The extent $N$ of the sliding window must therefore be limited, to maintain the advance
warning period required for safe crossing. 

Suppose that the sliding window falls either on a traffic gap,
modelling the Activity signal as  $A_n=\nu_n$, where $\nu(n)$ is random noise,
or at the precise moment a vehicle emerges, such that $A_n=s_n+\nu_n$, where
$s_n$ is the clean Activity pulse shape for the specific vehicle and viewing conditions.
The `TRAFFIC' and `GAP' state hypotheses are thus
\begin{equation}
\begin{array}{lllllll}
\theta_{1}: &  & A_n=s_n+\nu_n &  &  &  & A_n\sim\mathcal{N}[s_n,\sigma^{2}]\\
\theta_{0}: &  & A_n=v_n &  &  &  & A_n\sim\mathcal{N}[0,\sigma^{2}]
\end{array}
\end{equation}
where $\nu_n$ is modelled as zero-mean additive
white Gaussian noise.

The Activity values $A_n$ within a sliding window of length $N$ can be
represented as a Gaussian random vector $\mathbf{y}$, with the
likelihood functions
\begin{align}
\begin{array}{l}
f_\mathbf{y}(\mathbf{y}|\theta_{1})=\frac{1}{(2\pi)^{N/2}\sigma^{N}}\prod\limits _{n=0}^{N-1}\exp[-\frac{(A_{n}-s_{n})^{2}}{2\sigma^{2}}]\\
f_\mathbf{y}(\mathbf{y}|\theta_{0})=\frac{1}{(2\pi)^{N/2}\sigma^{N}}\prod\limits _{n=0}^{N-1}\exp[-\frac{A_{n}^{2}}{2\sigma^{2}}]
\end{array}
\end{align}
To discriminate between the two hypotheses we apply the Likelihood Ratio Test
\begin{equation}
\mbox{LRT}=\frac{\frac{1}{\left(2\pi\right)^{N/2}\sigma^{N}}\prod\limits _{n=0}^{N-1}\exp[-\frac{\left(A_{n}-s_{n}\right)^{2}}{2\sigma^{2}}]}{\frac{1}{\left(2\pi\right)^{N/2}\sigma^{N}}\prod\limits _{n=0}^{N-1}\exp[-\frac{A_{n}^{2}}{2\sigma^{2}}]}\gtrless\lambda
\end{equation}
where $\lambda$ is a discriminative threshold. This leads
to the detection rule:
\begin{equation}
\tilde{y} \equiv \sum\limits _{n=0}^{N-1}A_{n}s_{n}\gtrless\frac{1}{2}\left(2\sigma^{2}\ln\lambda+\sum\limits _{n=0}^{N-1}s_{n}^{2}\right) \equiv \gamma
\label{eq:detection}
\end{equation}
where $\tilde{y}$ denotes the correlation of the Activity signal with the pulse template.
The pulse shape depends on the particular installation, and can 
be learned during training. For a given installation, specific vehicle characteristics
lead to some variability.
The right hand side is an application-dependent threshold denoted $\gamma$.
We determine the threshold using the Neyman-Pearson criterion,
tuned to obtain a specified false-alarm probability.

Given either hypothesis, the random variable $\tilde{y}$ is Gaussian with variance 
$\textstyle{\sigma^{2}\sum\limits _{n=0}^{N-1}s_{n}^{2}}$. 
The probability of false alarm is determined by the tail distribution of the 
`GAP' hypothesis  $\theta_{0}$:

\begin{align}
P_{FA} & =\int\limits _{\gamma}^{\infty}f_{\tilde{y}}(z|\theta_{0})dz=Pr\left(\tilde{y}>\gamma|\theta_{0}\right) \nonumber \\ 
& =Pr\left(\frac{\tilde{y}}{\sigma\sqrt{\sum s_{n}^{2}}}>\frac{\gamma}{\sigma\sqrt{\sum s_{n}^{2}}}\right) \nonumber \\
&=Q\left(\frac{\gamma}{\sigma\sqrt{\sum s_{n}^{2}}}\right)=1-\phi\left(\frac{\gamma}{\sigma\sqrt{\sum s_{n}^{2}}}\right)
\end{align}
where $\phi(x)$ is the cumulative distribution function of the standard
normal distribution and $Q(x)$ is its tail probability, i.e.,  $Q(x)=1-Q(-x)=1-\phi(x)$.
The threshold 
$$
\gamma=Q^{-1}\left(P_{FA}\right)\sigma\sqrt{\sum s_{n}^{2}}
$$
thus depends on the admissible false-alarm rate, the noise and the Activity pulse template. The decision rule can be 
reformulated as
\begin{equation}
\frac{\sum\limits _{n=0}^{N-1}A_{n}s_{n}}{\sqrt{\sum\limits _{n=0}^{N-1}s_{n}^{2}}} \gtrless Q^{-1}\left(P_{FA}\right)\sigma
\end{equation}
which is the familiar matched filter result, 
correlating the input sequence with the Activity pulse template we wish to detect.
The corresponding probability of detection can be shown to be
\begin{equation}
P_{D} = Pr\left(\tilde{y}>\gamma|\theta_{1}\right) =
Q\left(Q^{-1}\left(P_{FA}\right)-\frac{\sqrt{\sum s_{n}^{2}}}{\sigma}\right) ~.
\end{equation}

So far, we have discussed the most demanding detection condition,
the earliest detection of an incoming vehicle.
The Activity pulse typically increases as the vehicle approaches;
thus, once the correlator output $\tilde{y}$ crosses the threshold, it generally exceeds the threshold 
as long as the vehicle has not reached the crossing.
When a vehicle appears late (e.g. leaving a parking space close to the crossing), the Activity pulse
rapidly ascends, and the correlator promptly crosses the threshold.

\subsection{Setting Detector Parameters}

The detector is tuned according to the noise variance and the Activity pulse template. 
During training, by the time the influx map is computed, the system produces 
Activity measurements. After sufficient training, the measurements are 
used to automatically estimate the Activity pulse template and the noise variance.

\begin{figure}[p]
\begin{center}
\includegraphics[width=17pc]{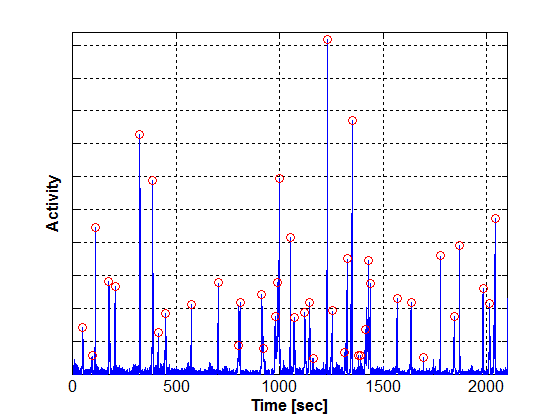}
\\[0.3cm]
\includegraphics[width=17pc]{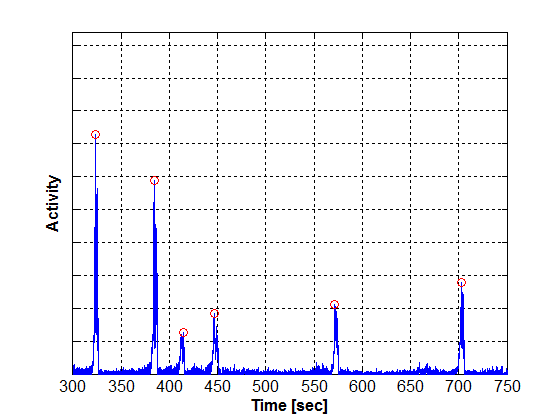}
\\[0.3cm]
\includegraphics[width=17pc]{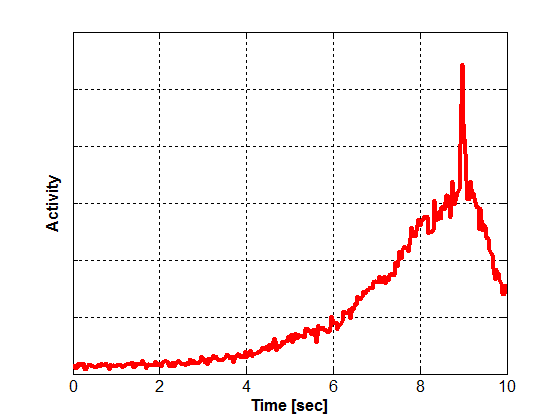}
\end{center}
\caption{Generating the Activity pulse template. \label{fig:template-generation}
Top: The Activity signal over a 35 minutes training period, 
with the salient local maxima detected.
Middle: 
Zoomed-in view of the Activity signal.
Bottom: The Activity pulse template, generated as the median of the Activity signal windows.
}
\end{figure}

The system locates the salient local maxima of the Activity signal during training, 
and sets a temporal window containing each maximum, where most of the window precedes the maximum. 
We set the Activity pulse template as the median of the Activity signal windows,
see Fig.~\ref{fig:template-generation}.

\section{Examples and Evaluation}

\begin{figure}[t]
\begin{center}
\includegraphics[width=20pc]{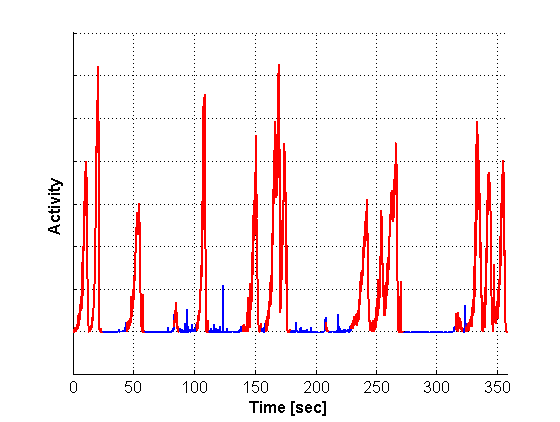}
\end{center}
\caption{The Activity signal computed over six minutes, including the events shown in 
Figs.~\protect\ref{fig:snaps-single-car}-\protect\ref{fig:snaps-several-cars}.
TRAFFIC and GAP indications are represented by red and blue coloring respectively. }
\label{fig:activity-signal}
\end{figure}

\begin{figure}[p]
\begin{center}
\includegraphics[width=22pc]{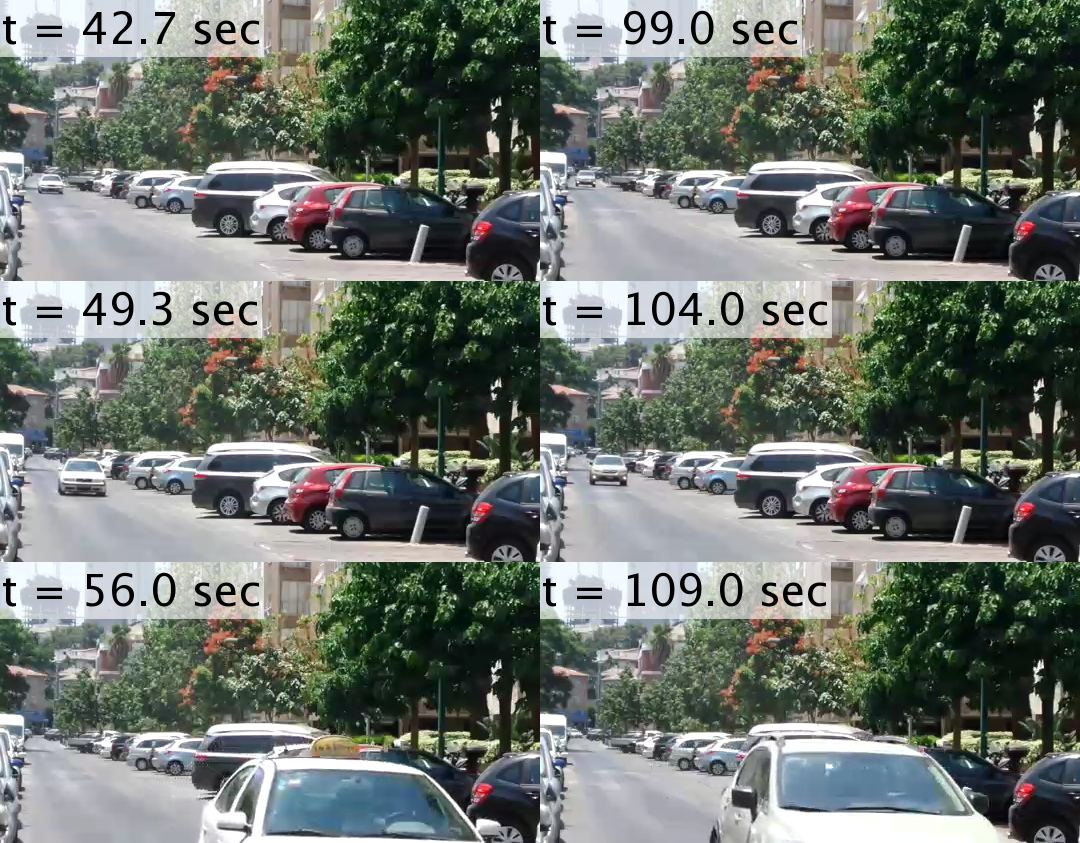}
\end{center}
\caption{Video frames corresponding to two single car events.}
\label{fig:snaps-single-car}
\end{figure}

\begin{figure}[p]
\begin{center}
\includegraphics[width=22pc]{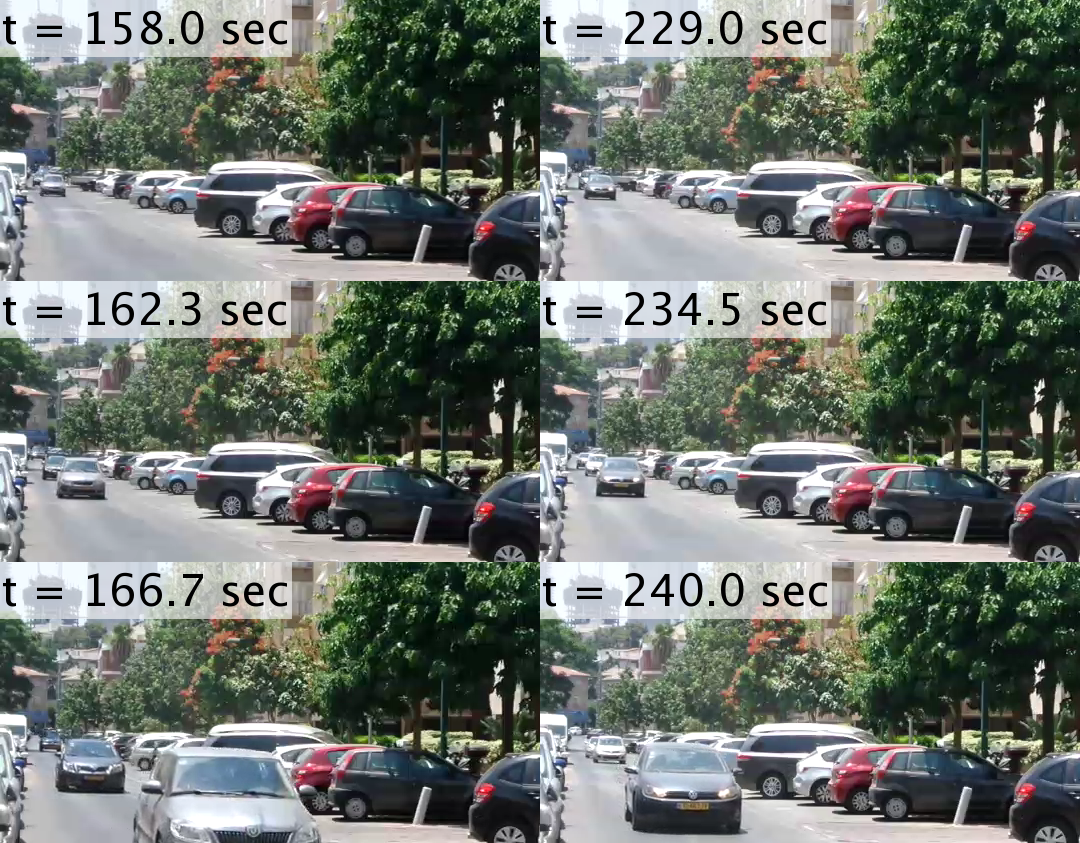}
\end{center}
\caption{Video frames corresponding to two multiple car events.}
\label{fig:snaps-several-cars}
\end{figure}

\begin{figure}[t!]
\begin{center}
\includegraphics[width=20pc]{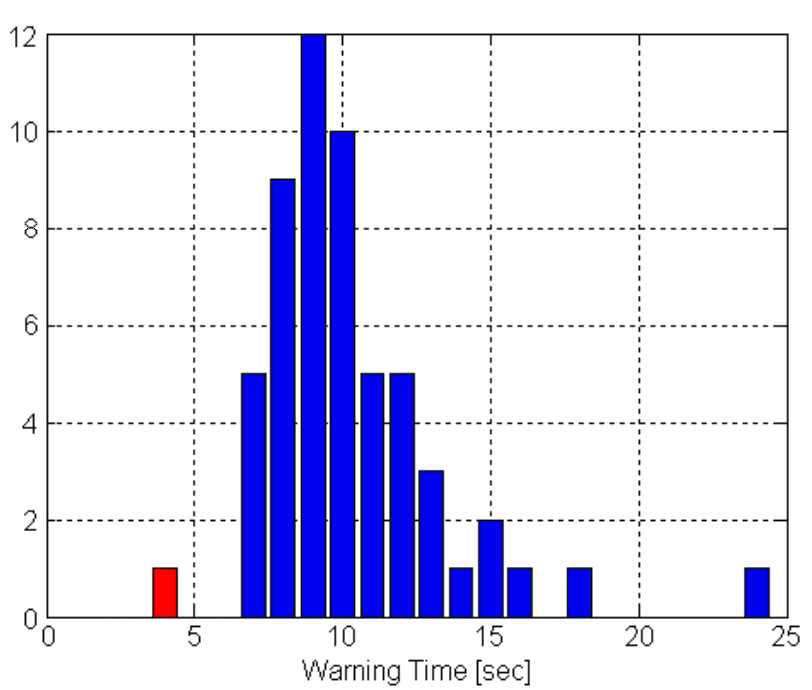}
\end{center}
\caption{Histogram of advance warning times provided by the real-time implementation during two hours
at the test site shown in Figs.~\protect\ref{fig:snaps-single-car}-\protect\ref{fig:snaps-several-cars}.}
\label{fig:warning_times_app}
\end{figure}

Fig.~\ref{fig:activity-signal} shows the activity signal computed during six minutes
at a test site. The color represents the prediction issued to the pedestrian,
red signifying TRAFFIC and blue meaning GAP.
TRAFFIC is typically indicated about 8-10 seconds
before the Activity signal peak, i.e. before the car reaches the camera, 
allowing secure crossing with sufficient margin.
Fig.~\ref{fig:snaps-single-car} presents selected frames from two approaching car 
events, corresponding to the unimodal activity pulses at  $t \approx 50s$ and at $t \approx 110s$. 
Multimodal activity pulses, e.g. at  $t \approx 160s$ and at  $t \approx 250s$ are each due to several
cars arriving in sequence. Corresponding snapshots are shown in Fig.~\ref{fig:snaps-several-cars}.
The false alarms at $t \approx 80s$ and $t \approx 205s$ are not due to random noise:
the first was caused by a person on the road, the second by a cat.
The random noise level is rather low, as seen circa $t=300s$ in Fig.~\ref{fig:activity-signal}. 
 Fig.~\ref{fig:warning_times_app} is the histogram of advance warning times provided by the system 
during two hours of testing.

\begin{figure}[p]
\begin{center}
\includegraphics[width=18pc]{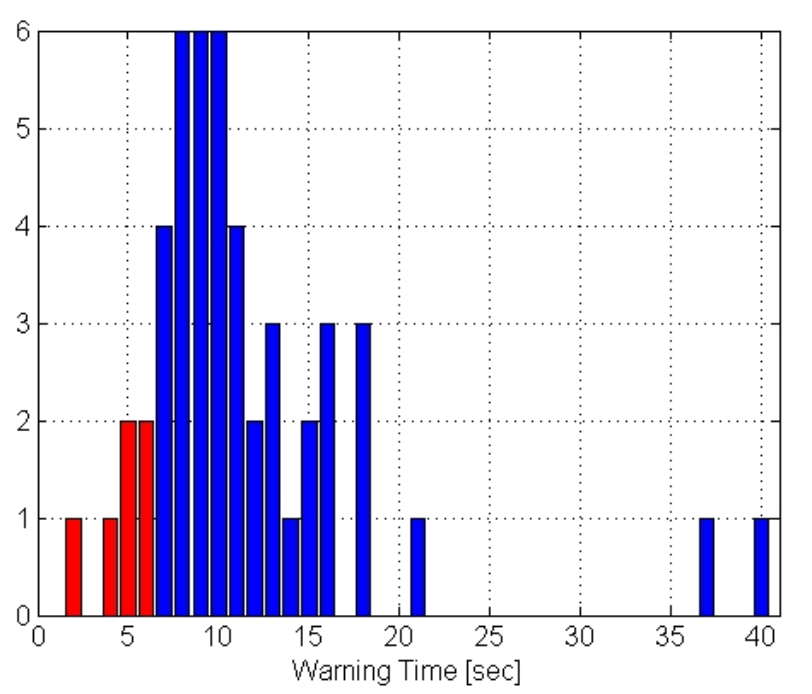}
\end{center}
\caption{Histogram of advance warning times obtained
during one hour at the test site shown in Fig.~\protect\ref{fig:snapshots-ex-all}.}
\label{fig:warning_times}
\end{figure}

\begin{figure}[p]
\begin{center}
\includegraphics[width=12pc]{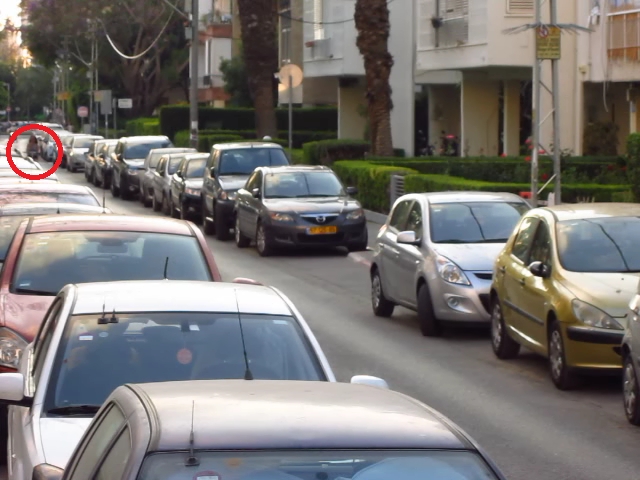} $~$
\includegraphics[width=12pc]{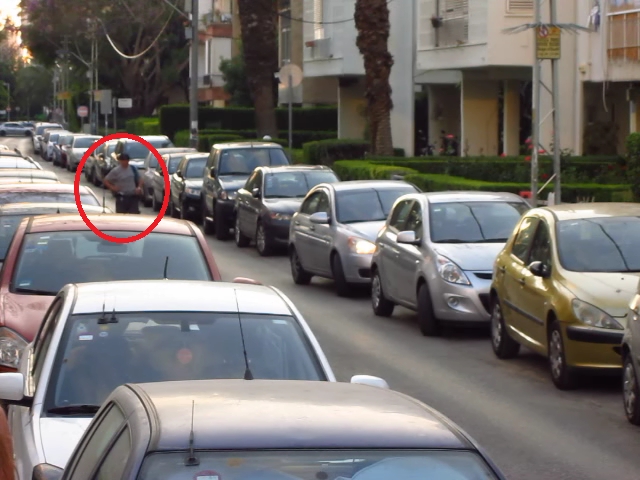}
\end{center}
\begin{center}
\includegraphics[width=12pc]{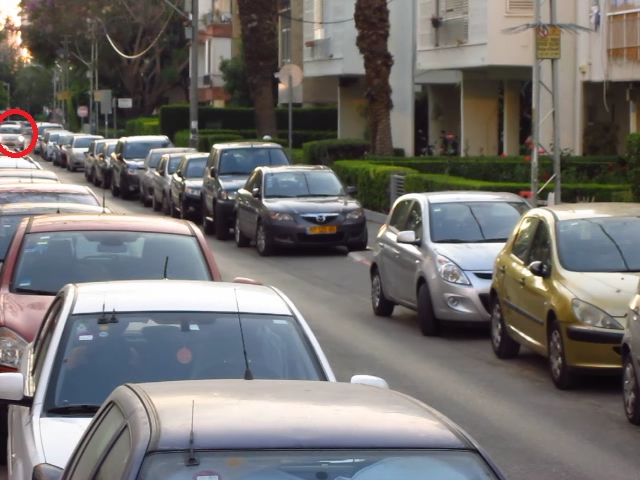}
\end{center}
\caption{Irregular events on the road. 
Top-left: A slow, distant bicyclist.
Top-right: A person suddenly appearing on the road.
Bottom: A car that appeared, stopped, and
eventually turned right into a side street.}
\label{fig:snapshots-ex-all}
\end{figure}

In the archetypal case, an approaching vehicle emerges 
close to the point of first appearance, thus maximizing the achievable warning time. 
However, in various urban or suburban settings, a 
vehicle can exit a parking place close to the camera, 
or turn into the monitored road from an adjacent driveway. 
Thus, while the typical visibility period of an incoming vehicle is often
greater than 10s, in unusual cases 
it can be as short as 1s, providing neither the system nor an alert human 
with sufficient warning.
Fig.~\ref{fig:warning_times} is the histogram of the advance 
warning times obtained during one hour at the challenging
installation shown in Fig.~\ref{fig:snapshots-ex-all}, characterized by
substantial occlusion near the point of first appearance and a residential afternoon 
activity pattern, including parking, biking and strolling.
Short warning times usually correspond to bikes and very slow vehicles,
that impose minimal hazard.
Irregular events acquired by the system, leading to unusually short or
long advance warnings, are exemplified in Fig.~\ref{fig:snapshots-ex-all}.
Additional results, including ROC curves, 
can be found in~\cite{Perry2014}.

\section{Real-Time Implementation}

To show that low-cost realization and wide deployment are possible, 
we implemented the system using an off-the-shelf Android device. 
Since the solution requires a narrow field-of-view lens,
we selected the Samsung Galaxy K-zoom smartphone, that has a built-in variable
zoom lens. Note that the variability of the optical zoom lens 
is not necessary, and the system can be readily built and operated using a 
fixed-zoom lens. Fixed-zoom lenses are widely available as low-cost 
smartphone accessories. Furthermore, unless remote management and
distributed cooperative operation are sought, the cellular communication 
parts of the smartphone are redundant. In addition, although the K-zoom
has a GPU capable of reducing computation time and increasing the frame rate,
our implementation only utilized the smartphone's CPU. 
Thus, the system can be implemented 
on extremely low-cost Android tablet-like devices. In many applications, the  
screen of the Android device is not in use, and can be shut down 
or removed altogether to further reduce the power consumption and cost.
 
\begin{figure}[t]
\begin{center}
\includegraphics[width=30pc]{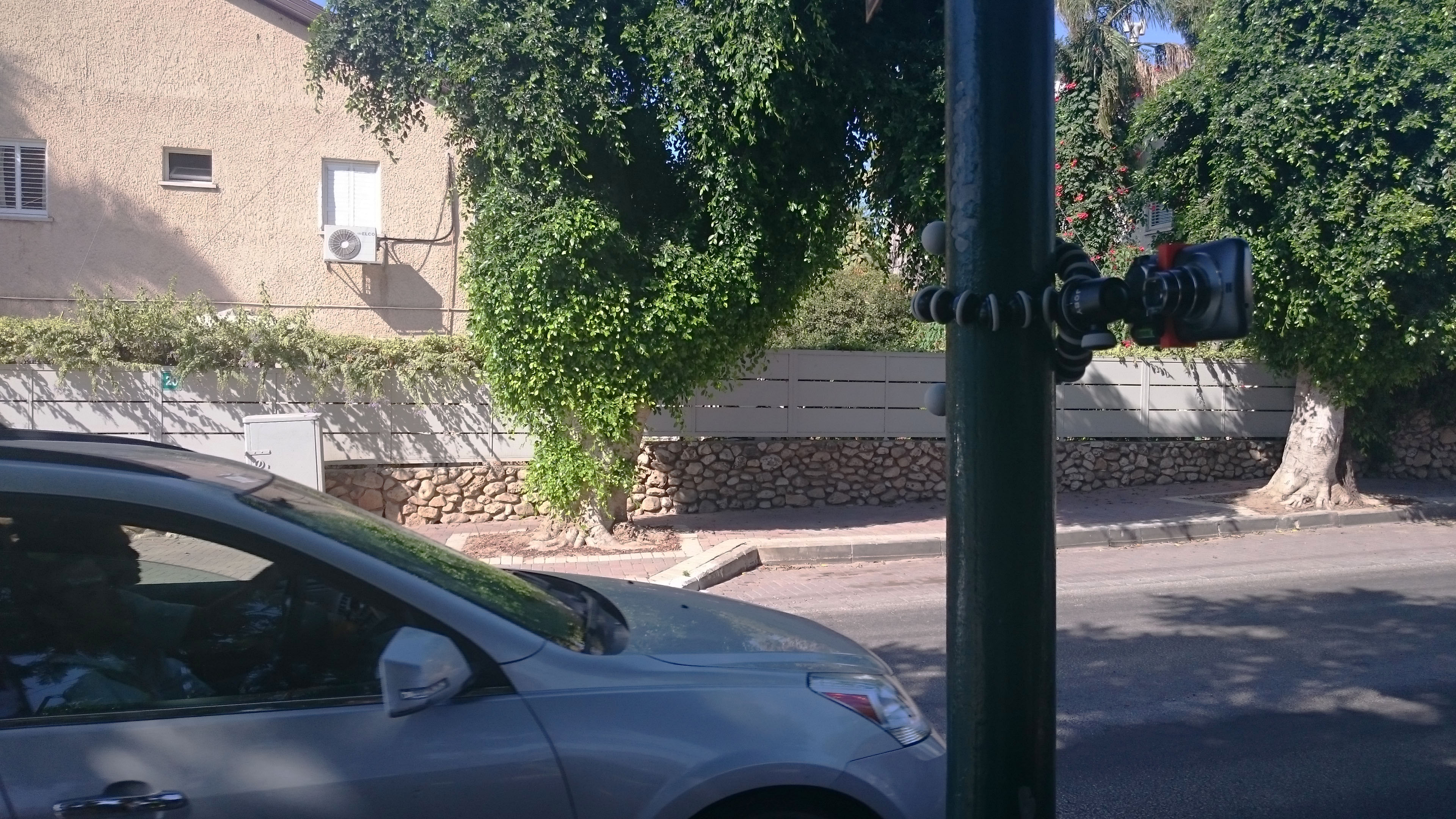}
\end{center}
\caption{The system, implemented on a Samsung Galaxy K-zoom smartphone,
attached to a lighting pole. Note the extended position of the zoom lens.} 
\label{fig:Zoom}
\end{figure}

Other than its variable zoom lens, the Samsung Galaxy K-zoom, introduced in 2014, is
similar to smartphones of its generation, with Android v4.4.2 (KitKat) 
as its native operating system.
Fig.~\ref{fig:Zoom} shows an installation of the system at a test site,
attached to a lighting pole, with its zoom lens extended.

The target system software was developed using OpenCV4Android SDK version 2.4.10, FFmpeg and JavaCPP. 
The most demanding computational task 
is optical flow estimation. During training, dense optical flow estimation is required for generating
the influx map. For that purpose, we used the OpenCV4Android implementation of 
Farneb\"{a}ck's algorithm~\cite{Farneback2003}, skipping frames to improve the estimation of the 
tiny optical flow vectors near the point of first appearance. Following the identification of that point,
a subset of 2000 image points was sampled using a Gaussian density function centered at the point of 
first appearance. In the online phase, sparse optical flow is evaluated only at those points, using the
OpenCV4Android implementation of the Lucas-Kanade~\cite{LucasKanade81} algorithm. 
The Activity signal is computed by
projecting the sparse optical flow field onto the corresponding sparse subset of the influx map.
A rate of 8 frames per second is easily obtained on the K-zoom smartphone. To reduce power 
consumption, the rate can be decreased to two frames per second without significant performance 
degradation.  
The template matching process~(\ref{eq:detection}) is carried out at 30 frames per second using temporal interpolation.
The results shown in Figs.~\ref{fig:activity-signal}-\ref{fig:warning_times_app} were obtained
using the real-time smartphone-based implementation.

Two way streets, as encountered for example in the test site shown in Fig.~\ref{fig:system}, raise two issues.
First, warning about traffic approaching the crossing from the opposite direction can be obtained by installing an 
additional instance of the system with its camera pointing in that direction. Cellular, WiFi or Bluetooth communication 
between the two systems allows issuing a mutually agreed indication to the pedestrian.
Second, in a two way road, as the camera monitors incoming traffic, distancing vehicles may be seen in the 
field of view. To distinguish incoming from outgoing traffic based on optical flow, one might compute 
vector field measures such as divergence. In practice, assuming that the camera is installed higher than 
typical vehicles, the distinction can be based on the sign of the vertical component of the optical flow vector:
incoming vehicles are associated with descending optical flow vectors. In the Android implementation,
upwards-pointing vectors in the influx map are simply nullified. Indeed, the system works well 
at the test site shown in Fig.~\ref{fig:system} using that method.    

\section{Discussion}

The global number of visually impaired people worldwide has been estimated to be 285 
millions~\cite{WHO}, and many more have age-related disabilities limiting their 
road-crossing skills. Reliable mobile personal assistive devices for road crossing without 
traffic lights are yet to be invented, but in any case are likely to be beyond the 
means of needy people worldwide. The goal of this research has been the development 
of an effective, easy to use, low cost solution. It is facilitated by the stationary, location-specific approach,
avoiding system ego-motion altogether. System installation is simple, as the location-dependent 
influx map and the Activity pulse template are automatically obtained by training. 

In this expanded version of~\cite{Perry2014}, the major additional accomplishment is the demonstration of 
real-time stand-alone operation using an off-the-shelf Android device.
The retail price of a suitable Android tablet, including all necessary components, is less than US\$ 40
at the time of writing, and the marginal cost of a fixed narrow field lens is negligible.
The mass-production manufacturing bill of materials, especially for a screen-less version
of the system, is therefore likely to be widely affordable.     

Interestingly,  in their future work section~\cite{Baker2005} imagined a
stationary solution for the unsignalized road crossing problem. Nevertheless, our approach 
is substantially more robust, computationally lighter and better exploits stationarity 
when compared to the tracking-based method applied by~\cite{Baker2005} in their robotic system.

\begin{figure}[t!]
\begin{center}
\includegraphics[width=30pc]{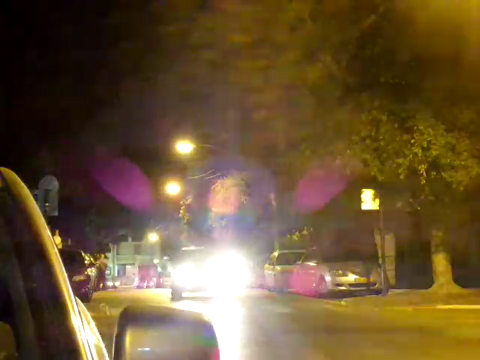}
\end{center}
\caption{Night-time imaging artifacts.} 
\label{fig:night}
\end{figure}

We evaluated the system at several test locations, obtaining promising results.
These experiments were carried out at fair weather conditions.
Additional work is needed to ensure reliable operation in the presence of rain or snow. 
Night-time operation is an additional challenge; the powerful headlights 
of an approaching car, aimed directly at the camera in the dark of night, create various 
imaging artifacts as seen in Fig.~\ref{fig:night}. Advance warning of an approaching vehicle 
after dark might be based on the presence of significant camera saturation.    

\bibliographystyle{abbrv}
\bibliography{main}
\end{document}